# Improving Cancer Hallmark Classification with BERT-based Deep Learning Approach

Sultan ZAVRAK[1*] and Seyhmus YILMAZ[1]

Sultan ZAVRAK

[1]Department of Computer Engineering, Duzce University, Duzce/TURKEY

sultanzavrak@duzce.edu.tr

***ORCID ID:*** 0000-0001-6950-8927

Seyhmus YILMAZ

[1]Department of Computer Engineering, Duzce University, Duzce/TURKEY

seyhmusyilmaz@duzce.edu.tr

***ORCID ID:*** 0000-0001-9987-2797

**<u>*Corresponding Author:</u>**

Sultan ZAVRAK

Department of Computer Engineering,
Faculty of Engineering, Duzce University, 81620, Duzce / TURKEY
Phone: +90 (380) 542 1036
E-mail: sultanzavrak@duzce.edu.tr

# Improving Cancer Hallmark Classification with BERT-based Deep Learning Approach

**Abstract**

This paper presents a novel approach to accurately classify the hallmarks of cancer, which is a crucial task in cancer research. Our proposed method utilizes the Bidirectional Encoder Representations from Transformers (BERT) architecture, which has shown exceptional performance in various downstream applications. By applying transfer learning, we fine-tuned the pre-trained BERT model on a small corpus of biomedical text documents related to cancer. The outcomes of our experimental investigations demonstrate that our approach attains a noteworthy accuracy of 94.45%, surpassing almost all prior findings with a substantial increase of at least 8.04% as reported in the literature. These findings highlight the effectiveness of our proposed model in accurately classifying and comprehending text documents for cancer research, thus contributing significantly to the field. As cancer remains one of the top ten leading causes of death globally, our approach holds great promise in advancing cancer research and improving patient outcomes.

**Keywords**: BERT, cancer hallmark classification, transfer learning, deep learning, natural language processing

## 1. Introduction

Cancer is one of the most difficult sicknesses for individuals in many parts of the world today, including epigenetic and genetic mutations (Jiang et al., 2020). Up to now, millions of people have died due to this disease in the world (Organization, 2008). The study of cancer has a long history that stretches from the past to the present and has consistently drawn the attention of biomedical researchers.

In the biomedical literature, cancer is frequently characterized by its hallmarks, a collection of related biological characteristics and behaviors that encourage the growth of cancer in the body (Baker, Korhonen, et al., 2016). The hallmarks of cancer are the features that are utilized to discriminate malignant cells from healthy cells (Yan & Wong, 2017). Scientists can take advantage of it to better comprehend how such cellular activities can cause cancer. These cellular procedures used to define how the disease grows are generalized by the hallmarks. In the seminal paper (Hanahan & Weinberg, 2000), the cancer hallmarks were first described by introducing six hallmarks, which were extended by four in a subsequent paper (Hanahan & Weinberg, 2011) to form the ten hallmarks that are currently recognized. Ten cancer hallmarks having a robust influence on cancer research and the discovery of its remedy have been explored until now : (i) sustaining proliferative signaling (PS), (ii) avoiding immune destruction (ID), (iii) evading growth suppressors(GS), (iv) activating invasion and metastasis (IM), (v) tumor-promoting inflammation (TPI), (vi) inducing angiogenesis (A), (vii) enabling replicative immortality (RI), (vii) resisting cell death (CD), (viii) deregulating cellular energetics (CE), (ix) genome instability and mutation (GI) (Baker, Silins, et al., 2016).

The ability to recognize cancer hallmarks is essential to the study of cancer and the investigation of its treatment. It is of the utmost importance, for the sake of the development of cancer research, to be able to predict whether or not a particular article or text contains specific information of interest by automatically classifying them into corresponding hallmarks.

The developments in Natural Language Processing (NLP) provide solutions to such issues. Text understanding and categorization, which includes the use of NLP techniques to recognize and classify biomedical text messages, is one method for classifying cancer hallmarks from biomedical text data. In recent years, the invention of transformers models in feature extraction (such as BERT) has significantly outperformed prior natural language frameworks in many text applications, such as sentiment analysis (Xu et al., 2019), spam detection (Yilmaz & Zavrak, 2022), etc. (Minaee et al., 2021) (Zhou et al., 2022), which offers an enormous opportunity to improve a robust classifier that can accurately classify text based on its hallmarks. In this context, a detection model based on state-of-the-art NLP methods, including BERT, for the text classification problem of recognizing cancer hallmarks in biomedical abstracts is developed. Consequently, one of our main goals in this study is to determine if BERT-based architectures can improve the performance of detecting hallmarks of cancer with insufficient training text data.

Recently, the utilization of pre-trained language models, such as BERT and ELMo, has demonstrated significant advancements in an array of text classification tasks, including sentiment analysis, emotion classification, and topic classification (Devlin et al., 2018)(Howard & Ruder, 2018)(Peters et al., 2018)(Radford et al., n.d.). The fundamental concept behind this approach is that a neural network language model, constructed from extensive corpora, offers a representation that amalgamates information from various levels of analysis, such as semantic, syntactic, morphological, and lexical (Barlas & Stamatatos, 2021). Consequently, these models can be fine-tuned on domain-specific corpora. In essence, this constitutes a form of transfer learning, as the knowledge acquired from one field is adapted and applied to accomplish tasks in another.

An important objective of this paper is to implement cutting-edge NLP techniques that can assist future researchers in overcoming problems and achieving a successful understanding of automatic solutions for the rapid and robust classification of cancer hallmarks. More specifically, this paper investigates biomedical text classification using deep learning (DL) instead of manual feature engineering. This study uses the problem setting and dataset of (Baker, Korhonen, et al., 2016), but instead of convolutional neural network (CNN), long short-term memory (LSTM), etc., it focuses on the pre-trained BERT model (Devlin et al., 2018) by employing transfer learning. BERT-based methods and "deep" networks are widely used in general NLP, but there has been little work applying them to biomedical text. In general, while DL and BERT-based techniques, in particular, are progressively common for general field NLP, there are comparatively few studies using this type of technique in biomedical text. In fact, to the best of our knowledge, there is no BERT-based model for hallmarks in the literature. Therefore, this study will be the first to use a BERT-based model for the detection of hallmarks of cancer for the first time. The most significant contributions to this article are summarized as follows:

- A biomedical text classification model is constructed on the BERT architecture using a corpus of over 1,800 biomedical abstracts annotated with ten hallmarks of cancer.
- The classification performances of the fine-tuned BERT models are evaluated and compared against previous text classification approaches.
- The experimental findings reveal that the proposed model exhibits a noteworthy improvement of at least 8.04% in comparison to prior text classification methodologies, as indicated by an accuracy score of 94.45%.

- These results confirm the effectiveness of the optimal architecture in effectively addressing the challenge of classifying cancer hallmarks in abstract texts.

This article is organized as follows. The related work in the field of biomedical text classification is carried out in Section 2. The proposed method used for the classification of the cancer hallmarks is explained in Section 3. Section 4 explains the publicly accessible hallmark dataset with an experimental setup and discusses the results of the models. In the last section, the concluding remarks are stated.

## 2. Related work

In this section, we will provide a synopsis of the research conducted in the field of biomedical text classification. Particularly, we will first discuss the traditional approaches to machine learning, which involve the manual creation of features, followed by research on deep neural architectures that employ feature learning rather than feature engineering.

### 2.1. Traditional methods

The appearance of a wide variety of biomedical text mining methods that concentrate on information discovery from scientific documents is born with the enormous volume of biomedical unstructured and structured data on cancer. Researchers have published many machine learning methods, including, Naive Bayes (Wang et al., 2007), Gradient boosting (Jiang et al., 2019), Maximum Entropy Modelling (Raychaudhuri et al., 2002), (Shatkay et al., 2008), Support Vector Machines (SVMs) (Cohen, 2006; Garla et al., 2013) etc. to develop the study on cancer. On the other hand, as will be shown in detail below, there are only a limited number of publications for cancer hallmark text classification up to now.

A dataset containing more than 1,800 abstracts from biomedical publications annotated with the ten hallmarks of cancer is introduced by Baker et al. (Baker, Silins, et al., 2016). In addition, they implemented a machine learning-based technique for categorizing the hallmarks of cancer. A traditional NL architecture by which they obtain a particular set of features utilized to feed SVM classifiers is employed by their technique. There are some different forms of features for example grammatical relations, verb classes, named entities, lemmatized bag-of-words, medical subject headings, chemical lists, and noun bi-grams. Decent results are obtained by detecting cancers with a mean F-score of 77%. On the other hand, the main drawback is the cost of including a long NL architecture with computational requirements.

Yan et al. concentrated on the curse of dimensionality and make a comparison among various cancer hallmark annotation techniques using 1580 PubMed abstracts (Yan & Wong, 2017). They implemented a new method called UDT-RF that utilizes ontological features. It used Medical Subject Headings (MeSH) ontology graph to enlarge the feature space. In addition, they use a variety of performance metrics to compare and evaluate state-of-the-art techniques, which reveal the complete performance spectrum on the full set of cancer hallmarks. To exhibit how the suggested method can show new insights into cancers, they carry out several case studies.

Though, manually generating a set of features takes a long time and consumes a lot of energy, and cannot guarantee enhanced results (Jiang et al., 2020). Although feature filtering approaches were used, detecting beneficial and significant features becomes more difficult as the size of the set is increased.

## 2.2. Deep learning methods

Nowadays, the approaches based on deep neural network (DNN) have accomplished advanced results in a variety of NLP applications for example machine translation, image categorization, semantic similarity, paraphrase recognition, language modeling, question answering, document summarization, and opinion mining (Jiang et al., 2020).

Researchers have begun to use DL in the biomedical area when DNNs accomplish great achievement in other domains, which includes relation extraction (Gao et al., 2019; Hanahan & Weinberg, 2000, 2011) concept extraction and coding (Garg et al., 2019), phenotyping (Hochreiter & Schmidhuber, 1997), named entity recognition (Bai et al., 2016; Lazebnik, 2010). Regarding cancer hallmark text classification, there is limited research that uses DNNs in the literature.

In (Baker, Korhonen, et al., 2016), the authors concentrated on classifying biomedical documents via DNN approaches that highlight feature learning instead of hand-crafted feature engineering. They used the dataset introduced by the authors of (Baker, Silins, et al., 2016) and implement also the same problem setting of (Baker, Silins, et al., 2016). Nonetheless, they emphasize CNNs rather than SVMs. The experiment results show that some CNN initialization approaches, training procedure, and hyper-parameters can permit the network to accomplish better results than a previously suggested SVMs model that use hand-built engineered features containing named entity recognition and syntactic analyses outputs.

To utilize label co-occurrence relations for example hypernymy, the authors of (Baker & Korhonen, 2017) implemented a novel technique for hierarchical multi-label document classification that initializes a last hidden layer of a base CNN framework on the network of Kim (Kim, 2014), which is commonly employed in text classification applications. They explore their method with two multi-label classification problems in the biomedical area utilizing both document and sentence level classification: In the experiment setup, this technique accomplished encouraging performance.

A new model called Deep Contextualized Attentional Bidirectional LSTM (DECAB-LSTM) for the cancer hallmark text classification is implemented by Jiang et al. (Jiang et al., 2020). This method uses a contextual attention mechanism to able to learn to capture the most significant part of a sentence. In addition, the impact of a decent word embedding for the cancer hallmark text classification is explored. In a supervised learning environment, some experiments are done on a benchmark dataset. They utilize several evaluation metrics such as accuracy, macro-F1 score, and AUC scores to test their methods by using ten hallmarks of cancer. The advanced results over the baseline architectures for the cancer hallmark text classification are achieved by the model during the experiment.

The BERT (Bidirectional Encoder Representations from Transformers) is a pre-trained language model that has revolutionized natural language processing (NLP) tasks by achieving state-of-the-art performance in a variety of text classification tasks. Therefore, by using BERT architecture, it is possible to achieve better results in classification of cancer hallmarks than the traditional and deep learning methods mentioned in the previous studies. Additionally, the BERT model can provide a more accurate and robust classification of cancer hallmarks due to its ability to capture the complex and nuanced relationships between different concepts in biomedical text data.

## 3. The proposed model architecture

This study leverages the BERT model, a state-of-the-art bidirectional transformers architecture that has demonstrated impressive performance across a range of natural language processing (NLP) applications (Qasim et al., 2022).

### 3.1. BERT

The BERT network architecture, originally introduced by Devlin et al. (Devlin et al., 2018), is a multi-layer bidirectional transformer encoder that is based on the original model proposed by the authors of (Vaswani et al., n.d.). The BERT (Devlin et al., 2018) is a straightforward yet strong model and, achieves considerable improvement in numerous machine learning applications. Training bidirectional representations from an unlabeled dataset are the primary objective of BERT, which can operate on collaborative left and right context phenomena across all layers (Qasim et al., 2022).

Pre-training and fine-tuning are two different steps used by BERT (Devlin et al., 2018). In the stage of pre-training, unlabeled data is used to train the network over various pre-training tasks. Firstly, the initialization of the BERT network is done by using the pre-trained parameters, and fine-tuning is applied to the entire parameters utilizing a labeled dataset, which is taken from the downstream NLP tasks. Although they are started using the same pre-trained parameters, there are distinct fine-tuned networks in each downstream problem. A distinguishing feature of BERT is its unified model across dissimilar downstream applications (Devlin et al., 2018). The distinction is minimum between the last downstream model and the pre-trained model.

In this study, we use the BERT-base model and the quantity of self-attention heads is shown as A, the hidden size as H, and the number of layers (i.e., Transformer blocks) as L in the model. The framework of the BERT model in the task of classification of hallmarks is illustrated in Figure 1.

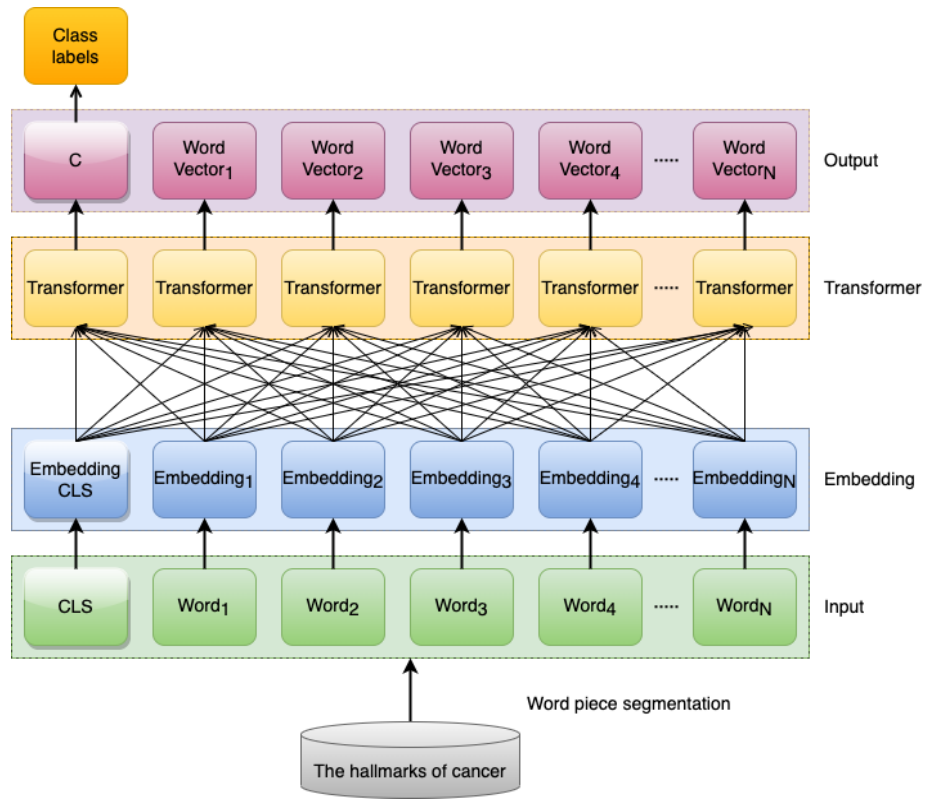

Figure 1. The framework of the BERT model in the hallmark classification task (Xiaofeng et al., 2021)

Initially, the transformer is introduced as a feature extractor to advance the performance of translation tasks. Later on, researchers prove that after inventing a self-attention mechanism and positional encoding, the transformer can be efficiently used in various artificial intelligence areas as well. An encoder-decoder structure is used by transformers. The decoder and encoder are architecturally identical, and they both contain six similar layers. An additional attention sublayer is present in the decoder but not in the encoder. There are two sublayers in each layer: a fully connected feedforward neural network layer and a multiheader self-attention layer (Xiaofeng et al., 2021). The architecture of the encoder is illustrated in Figure 2.

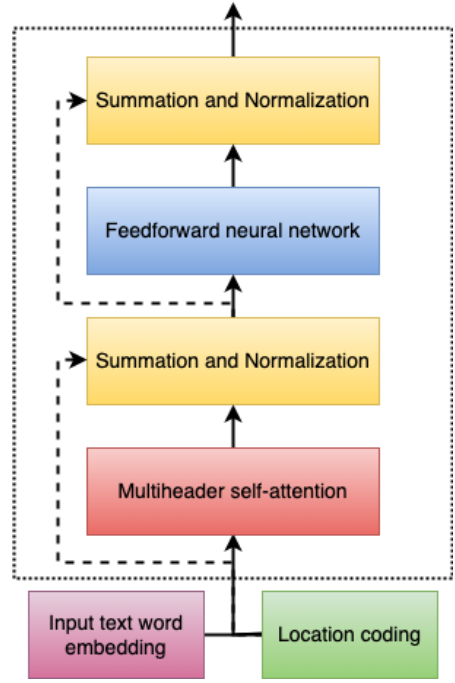

Figure 2. The architecture of the Encoder (Xiaofeng et al., 2021)

The multiheader self-attention layer, which is a combination of some self-attention architectures is one of the most significant components in transformers. Both encoders and decoders have the multiheader self-attention layer. The decoder takes care of the portions of the text documents required to be translated while training the network when the attention mechanism was first announced. To predict the output, the dissimilar inputs' weight is computed according to the attentional mechanism. The self-attention mechanism calculates attention between the target text and the original text. The attention mechanism, on the other hand, calculates the attention between the target text and the original text. This improves the model's ability to determine the characteristics between distant sentence elements. In this way, it permits the network to capture the features between more distant elements in the text. The formula to compute the matrix for the attention mechanism is shown in Eq. 1 (Xiaofeng et al., 2021).

$$Attention(Q, K, V) = softmax\left(\frac{QK^T}{\sqrt{d_k}}\right)V \quad\quad 1$$

According to the above equation, V represents the value matrix, K represents the key matrix and Q represents the query matrix. Such values are the product of the matrix acquired by the text embedding and their respective weight matrices. A query vector dimension is represented by $d_k$. Based on this, more advanced performance is obtained by the multiheader self-attention mechanism in comparison with the single self-attention mechanism. To increase the ability of the network to concentrate on words in dissimilar places, key weight matrices, some query weight matrices, and value weight matrices are utilized. The equation is calculated as shown in Eq. 2 (Vaswani et al., n.d.; Xiaofeng et al., 2021):

$$MultiHead(Q, K, V) = Concat(head_1, \ldots, head_h)W^O \quad\quad 2$$

According to the equation above, $W^O$ shows the weight matrix of the joint training and the projection matrix of the weight matrix acquired via the $i$-th attention head are $head_i =$

$Attention(QW_i^Q, KW_i^K, VW_i^V)$ , $W_i^Q$, $W_i^K$ and, $W_i^V$. To specify the encoding of the present word, the self-attention mechanism allows the network to encode a word by obtaining other words in the text as reference factors in the NLP processing (Vaswani et al., n.d.). The absolute/relative position order knowledge of the input is not stated by the self-attention mechanism alone, although the positional knowledge is very significant for every token while processing text documents (Xiaofeng et al., 2021). For that reason, a vector is put into the embedding of each input token in transformers to specify its position knowledge such as positional encoding to resolve this challenge. The formula to be calculated is shown as shown in Eq. 3 and 4.

$$PE_{(pos,2i)} = \sin\left(pos/10000^{2i/d_{\text{model}}}\right) \quad\quad 3$$

$$PE_{(pos,2i+1)} = \cos\left(pos/10000^{2i/d_{\text{model}}}\right) \quad\quad 4$$

In Eq. 3 and 4, $i$ is the dimension and $pos$ is the positional knowledge of the token in the text. If $pos = 0,1,2,\ldots,L-1$ and $L$ are taken as the length of the sentence, the vector length in the word model is denoted by $dmodel$, the $i$-th dimension is shown by $i$ in the positional encoding vector, thus $i = 0,1,2,\ldots,(dmodel/2 - 1)$. Therefore, the transformer in the BERT architecture utilizes positional embedding to show positional knowledge in text documents and advances the original positional encoding (Xiaofeng et al., 2021). In addition to this, to capture as much contextual data as possible in the texts when training, BERT consists of a bidirectional architecture.

The output vectors of BERT are implemented while utilizing BERT for word embedding. There are 12 transformer encoder blocks in BERT-base architecture (Park et al., 2022). Feed-forward network, normalization, and self-attention components are included in a transformer encoder block. A series of words tokenized utilizing a byte-pair encoding (BPE) tokenizer is input $X$. The correlation between words is represented by $Z$ in input $X$. The weighted sum and the scaled dot-product attention mechanism are used to compute $Z$. The self-attention mechanism is represented by the equations from Eq. 5 to Eq. 11 (Park et al., 2022).

$$X = [x_1, x_2, \cdots, x_n] \quad\quad 5$$

$$Q = W_q \times X = [q_1, q_2, \cdots, q_n] \quad\quad 6$$

$$K = W_k \times X = [k_1, k_2, \cdots, k_n] \quad\quad 7$$

$$V = W_v \times X = [v_1, v_2, \cdots, v_n] \quad\quad 8$$

$$Z = softmax\left(\frac{QK^t}{\sqrt{d_k}}\right)V = [z_1, z_2, \cdots, z_n] \quad\quad 9$$

$$\dot{Z} = Normalize\ (X + Z) \quad\quad 10$$

$$\ddot{Z} = Normalize\ \left(X + FFN(\dot{Z})\right) \quad\quad 11$$

As formulated in Eq. 12 and 13, the normalization and feed-forward steps are used to compute the last vector $\ddot{Z}$. The $\ddot{Z}$ of the final layer was utilized as the word embedding after the stage described above was done a few times. Every embedding vector is dependent on the text containing the word. For that reason, these sorts of vectors are called context embeddings.

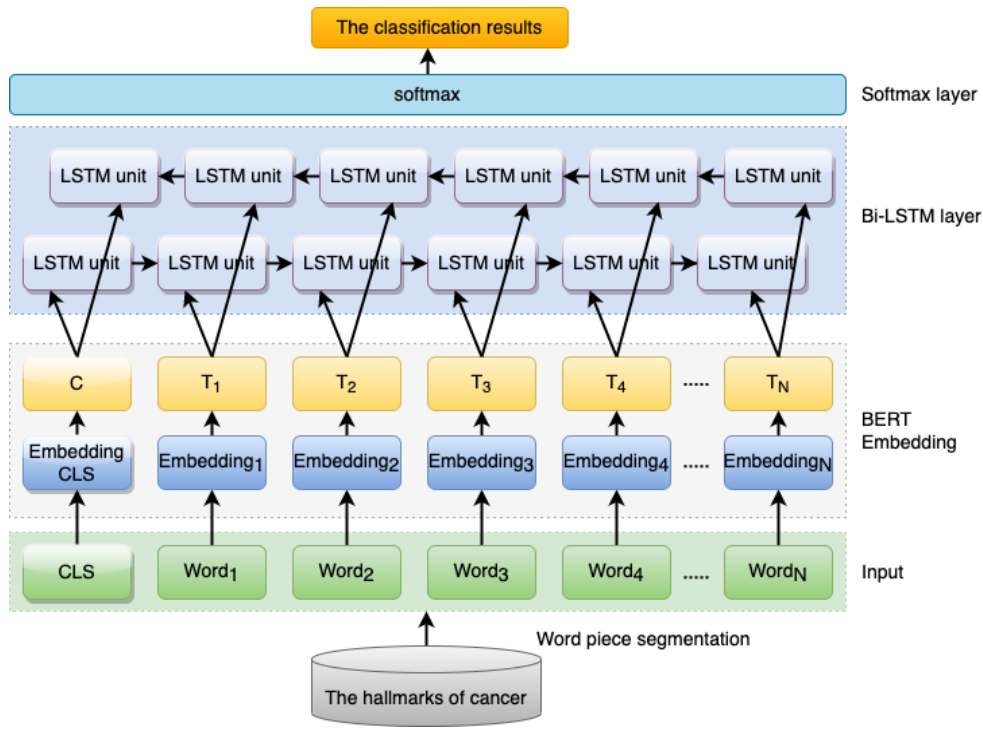

Figure 3. The diagram of the model structure

Pre-training and fine-tuning are two different steps used by BERT (Devlin et al., 2018). In the stage of pre-training, unlabeled data is used to train the network over various pre-training tasks. Firstly, the initialization of the BERT network is done by using the pre-trained parameters, and fine-tuning is applied to the entire parameters utilizing a labeled dataset, which is taken from the downstream NLP tasks. Although they are started using the same pre-trained parameters, there are distinct fine-tuned networks in each downstream problem.

In BERT, fine-tuning is a sort of supervised learning (Park et al., 2022) and it is used in some domains for example text classification. In general, fine-tuning uses [CLS], which is a special symbol input put into the start of every input sample (Devlin et al., 2018), when BERT is applied to some tasks such as sentiment analysis, and question. the [CLS] representation is used to compute the classification likelihood for the label. Pooling is applied to the last hidden state of the CLS symbol. By employing the SoftMax function via the classification layer, the label with the maximum likelihood is computed. To estimate the log-likelihood of the accurate label, the classification layer and the pre-trained BERT parameters are fine-tuned afterward.

## 4. Experiments

### 4.1. Dataset Description

The present study employs a corpus of 1852 biomedical publication abstracts that have been annotated with the hallmarks of cancer, as documented in the literature (Baker, Silins, et al., 2016). In this study, the proposed models are trained and assessed using this corpus. Each abstract in the dataset possesses the potential to be associated with none or more of the ten hallmarks under scrutiny. Consequently, the objective of the study pertains to multi-label classification, an approach explicated in greater detail below (Zhou et al., 2022).

**Sustaining proliferative signaling**: To develop and split, normal cells need molecules to act as signals. In contrast to normal cells, malignant cells can develop without such signals.

**Evading growth suppressors:** There are stages in cells, which stop development and separation. Such stages in cancer cells are changed so that they do not efficiently stop cells from separating.

**Resisting cell death:** There are mechanisms called apoptosis in which cells are automatically dead when they are injured. Such a mechanism can be overcome by malignant cells.

**Enabling replicative immortality:** When a particular number of separations occur, normal cells are dead. But malignant cells can grow and divide infinitely, which is called immortality.

**Inducing angiogenesis:** Fresh blood vessels are produced via a process called angiogenesis can be started by malignant cells. Therefore, the necessary oxygen and other nutrients are provided.

**Activating invasion & metastasis:** To occupy neighboring textures and increase the spread of areas of the body distant, malignant cells can leave their initial place.

**Genome instability & mutation:** Heavy chromosomal abnormalities, which get worse as the sickness grows, usually can be in malignant cells.

**Avoiding immune destruction:** The immune system is not able to see malignant cells.

**Deregulating cellular energetics:** Aberrant metabolic pathways are used by many malignant cells to produce energy, for example generating glucose fermentation even if there is sufficient oxygen to appropriately inhale.

**Tumor-promoting inflammation:** Inflammation contributes to the propagation, metastasis, and survival of malignant cells by influencing the microenvironment nearby tumors.

Table 1. Annotation statistics of the dataset

| Hallmark | Train (70.36%) | | Validation (9.88%) | | Test (19.76%) | | Total | |
|---|---|---|---|---|---|---|---|---|
| | pos | neg | pos | neg | pos | neg | pos | neg |
| Sustaining proliferative signaling | 328 | 975 | 43 | 140 | 91 | 275 | 462 | 1390 |
| Evading growth suppressors | 172 | 1131 | 22 | 161 | 46 | 320 | 240 | 1612 |
| Resisting cell death | 303 | 1000 | 42 | 141 | 84 | 282 | 429 | 1423 |
| Enabling replicative immortality | 81 | 1222 | 11 | 172 | 23 | 343 | 115 | 1737 |
| Inducing angiogenesis | 99 | 1204 | 13 | 170 | 31 | 335 | 143 | 1709 |
| Activating invasion and metastasis | 208 | 1095 | 29 | 154 | 54 | 312 | 291 | 1561 |
| Genomic instability and mutation | 227 | 1076 | 38 | 145 | 68 | 298 | 333 | 1519 |
| Tumor promoting inflammation | 169 | 1134 | 24 | 159 | 47 | 319 | 240 | 1612 |
| Cellular energetics | 74 | 1229 | 10 | 173 | 21 | 345 | 105 | 1747 |
| Avoiding immune destruction | 77 | 1226 | 10 | 173 | 21 | 345 | 108 | 1744 |

## 4.2. Experimental Setup

In the realm of supervised learning, a series of rigorous experiments were conducted on a benchmark dataset, wherein each text sequence was subjected to manual tagging to assign a hallmark. The training split was employed to train our model, and the most promising model was chosen using the development (validation) split. Subsequently, the test split was utilized to comprehensively evaluate the efficacy and performance of the model.

As a research platform, the present study employs Google Colab Pro, which provides a dedicated graphical processing environment based on Tesla P100 GPUs for experimentation. The ktrain (Maiya, 2022), a lightweight wrapper library for TensorFlow Keras, is utilized for training and testing deep learning and machine learning models.

For the classification of the hallmarks of cancer, the pre-trained BERT-base was initially loaded with relatively fewer parameters to run atop Keras with TensorFlow as the backend. In this study, both cased and uncased configurations of the BERT-base pre-trained model are utilized. The base model consists of 12 layers of transformer blocks, the hidden size is 768, the number of self-attention heads is 12, and contains 110 million parameters in total. The maximum length limit of tokens is set to 512. The batch size is set to 6 to provide small training sets at once. 1 Cycle policy (Smith, 2018) is utilized in training policy to determine optimal values for the tightly coupled learning rate, momentum, and regularization. We initiate the fit function for 20 epochs while keeping the learning rate at 1e-5.

### 4.3. Results and Discussion

Owing to a significant disparity in sample distribution, the proposed models have been evaluated based on macro-F1, macro-precision, macro-recall, and classification accuracy metrics, alongside the AUC score, across ten hallmarks of cancer (Song et al., 2022). The assessment results of these models are presented in Table 2, Table 3, and Table 4.

Table 2. The results of the proposed BERT-base-uncased model after finetuning (%)

| Hallmark | Accuracy | Macro-Precision | Macro-Recall | Macro-F1 | AUC |
|---|---|---|---|---|---|
| Sustaining proliferative signaling | 93.44 | 85.36 | 91.56 | 88.04 | 97.17 |
| Evading growth suppressors | 97.81 | 89.90 | 89.90 | 89.90 | 99.50 |
| Resisting cell death | 99.18 | 93.75 | 99.57 | 96.45 | 99.79 |
| Enabling replicative immortality | 99.18 | 97.44 | 95.51 | 96.45 | 98.11 |
| Inducing angiogenesis | 90.98 | 78.92 | 84.61 | 81.36 | 93.64 |
| Activating invasion and metastasis | 92.90 | 87.71 | 89.39 | 88.52 | 94.85 |
| Genomic instability and mutation | 95.36 | 82.87 | 93.07 | 87.07 | 94.68 |
| Tumor promoting inflammation | 95.63 | 92.34 | 96.33 | 94.11 | 97.32 |
| Cellular energetics | 81.97 | 76.06 | 79.18 | 77.30 | 88.62 |
| Avoiding immune destruction | 96.99 | 92.37 | 94.65 | 93.46 | 97.19 |
| **Average** | **94.34** | **87.67** | **91.38** | **89.26** | **96.09** |

Table 2 presents the results of the fine-tuned BERT-base-uncased model, indicating accuracy values ranging from 81.97% to 99.18% across individual classification tasks. Similarly, Table 3 displays the outcomes of the fine-tuned BERT-base-cased model, revealing accuracy scores ranging from 88.29% to 99.82% across individual classification tasks. The BERT-base uncased model exhibited superior performance than the BERT-base cased model, as evidenced by higher recall, F1, and AUC scores on average. Conversely, the BERT-base cased model demonstrated superior performance over the BERT-base uncased model with respect to accuracy and precision scores.

Table 3. The results of the proposed BERT-base-cased model after finetuning (%)

| Hallmark | Accuracy | Macro-Precision | Macro-Recall | Macro-F1 | AUC |
|---|---|---|---|---|---|
| Sustaining proliferative signaling | 93.44 | 85.36 | 91.56 | 88.04 | 96.97 |
| Evading growth suppressors | 98.09 | 91.92 | 90.04 | 90.96 | 98.87 |
| Resisting cell death | 99.18 | 93.75 | 99.57 | 96.45 | 99.82 |
| Enabling replicative immortality | 99.18 | 95.69 | 97.53 | 96.59 | 98.44 |
| Inducing angiogenesis | 90.98 | 79.57 | 79.02 | 79.29 | 89.36 |
| Activating invasion and metastasis | 92.62 | 87.96 | 87.52 | 87.74 | 95.96 |
| Genomic instability and mutation | 96.17 | 85.88 | 92.06 | 88.65 | 96.02 |
| Tumor promoting inflammation | 94.81 | 91.77 | 94.12 | 92.87 | 97.01 |
| Cellular energetics | 83.61 | 78.00 | 81.00 | 79.23 | 88.29 |
| Avoiding immune destruction | 96.45 | 91.76 | 92.52 | 92.14 | 96.05 |
| **Average** | 94.45 | 88.17 | 90.49 | 89.20 | 95.68 |

Table 4 presents a comparison between the performance of various methods used in previous studies and the Fine-tuned BERT models in terms of accuracy, precision, recall, F1, and AUC. The results indicate that the fine-tuned BERT models outperform the previous studies in all metrics except AUC. The fine-tuned BERT-base-uncased model achieved an accuracy of 94.34%, which is substantially higher than the best accuracy reported in previous studies (86.3% for DECAB-LSTM (Jiang et al., 2020)). Similarly, the fine-tuned BERT-base-cased model achieved an accuracy of 94.45%, which is also higher than the best accuracy achieved in previous studies.

Table 4. Comparison with previous studies (%)

| Ref. | Method | Accuracy | Macro-Precision | Macro-Recall | Macro-F1 | AUC |
|---|---|---|---|---|---|---|
| Baker, Korhonen, and Pyysalo (Baker, Korhonen, et al., 2016) | CNN tuned | - | - | - | 81.0 | 97.6 |
| Jiang et al. (Jiang et al., 2020) | DECAB-LSTM | 86.3 | - | - | 89.10 | 98.9 |
| Prabhakar and Won (Prabhakar & Won, 2021) | Hybrid BiGRU with multihead attention | 74.71 | 70.82 | 68.99 | 69.89 | - |
| *This study* | Finetuned BERT-base-uncased | 94.34 | 87.67 | 91.38 | 89.26 | 96.09 |
| | Finetuned BERT-base-cased | 94.45 | 88.17 | 90.49 | 89.20 | 95.68 |

The superiority of the fine-tuned BERT models can be attributed to their extensive pre-training on text data, which results in a greater comprehension of language and its subtle nuances. Furthermore, BERT models employ bidirectional transformers, enabling them to capture the relationships between words in both directions, which is not possible with conventional models such as CNNs and LSTMs. Besides, using biomedical text data to fine-tune these models improves their understanding of this context, allowing them to extract more context and meaning from text data, which is essential for accurately classifying the hallmarks cancer.

## 5. Conclusion

In conclusion, this paper presents a novel approach for classifying biomedical text using BERT, a bidirectional transformer model. This method is evaluated using the hallmarks of cancer, which have become extremely important in cancer research, and the results of the experiments demonstrate that the proposed method achieves an accuracy of 94.45%, which is superior to almost all of the results that are currently available in the literature.

The contributions of this work are significant because while DNN-based methods are becoming more prevalent in a variety of NLP domains, there have been relatively few studies applying this strategy to biomedical text, specifically the cancer hallmark. The proposed model fills this gap and demonstrates the potential of using deep learning for cancer research.

Although the proposed model achieves a high level of accuracy, there are still some limitations to this work. Firstly, the proposed model only focuses on the classification of cancer hallmarks, and further research is required to investigate its performance on other types of biomedical text. Secondly, the size of the dataset used in this study is relatively small, and it may not be representative of all cancer types.

Future work can address these limitations by considering larger and more diverse datasets and exploring the performance of the proposed model on other types of biomedical text. In addition, further research can investigate the potential of combining different deep learning techniques to improve the accuracy of the proposed model.

**Data Availability**

The dataset used in this study was obtained from previously published works.

**Competing interests**

The authors declare that they have no conflict of interest.

**Funding**

Not applicable

**Authors' contributions**

Each author contributed equally to this study's Conceptualization, Methodology, Analysis, and Writing.